# Multi-scale Efficient Graph-Transformer for Whole Slide Image Classification

*Saisai Ding, Juncheng Li, Jun Wang, Member, IEEE, Shihui Ying, Member, IEEE, Jun Shi, Member, IEEE*

*Abstract*—The multi-scale information among the whole slide images (WSIs) is essential for cancer diagnosis. Although the existing multi-scale vision Transformer has shown its effectiveness for learning multi-scale image representation, it still cannot work well on the gigapixel WSIs due to their extremely large image sizes. To this end, we propose a novel Multi-scale Efficient Graph-Transformer (MEGT) framework for WSI classification. The key idea of MEGT is to adopt two independent Efficient Graph-based Transformer (EGT) branches to process the low-resolution and high-resolution patch embeddings (i.e., tokens in a Transformer) of WSIs, respectively, and then fuse these tokens via a multi-scale feature fusion module (MFFM). Specifically, we design an EGT to efficiently learn the local-global information of patch tokens, which integrates the graph representation into Transformer to capture spatial-related information of WSIs. Meanwhile, we propose a novel MFFM to alleviate the semantic gap among different resolution patches during feature fusion, which creates a non-patch token for each branch as an agent to exchange information with another branch by cross-attention. In addition, to expedite network training, a novel token pruning module is developed in EGT to reduce the redundant tokens. Extensive experiments on TCGA-RCC and CAMELYON16 datasets demonstrate the effectiveness of the proposed MEGT.

*Index Terms*—Whole slide images, cancer diagnosis, graph-Transformer, multi-scale feature fusion, cross-attention.

## I. INTRODUCTION

HISTOPATHOLOGICAL images are considered as the "gold standard" for cancer diagnosis, and the analysis of Whole Slide Images (WSIs) is a critical approach for cancer diagnosis, prognosis, survival prediction, and estimation of response-to-treatment in patients [1][2][3]. Nowadays, deep learning (DL) has been successfully applied to the field of computational pathology to develop the computer-aided diagnosis (CAD) system, which can help pathologists improve diagnosis accuracy together with good consistency and reproducibility [4][5].

Due to the huge size of WSIs and lack of pixle-level annotation, the weakly-supervised learning framework is generally adopted for analysis of gigapixel WSI, among which multiple instance learning (MIL) is a typical method [6][7]. Under the MIL framework, each WSI is regarded as a bag with numerous cropped patches as instances. Their features are then extracted and aggregated to produce a slide-level representation for the following weakly-supervised task. Existing MIL-based methods have shown effectiveness in WSI analysis [8][9][10][11][12]. However, these works generally focus on the single resolution of WSIs, and ignore the multi-resolution information. In fact, a WSI can be scaled at different resolutions and express rich multi-scale diagnostic information from extremely small cells to large tissues [13]. These multi-scale semantics can cover tumor-related information more comprehensively, thus helping pathologists improve diagnostic accuracy.

Inspired by the diagnosis process of pathologists, several works have extended the previous MIL frameworks to the multi-resolution oriented approaches [13][14][15][16]. For example, a simple strategy is to concatenate the instances from different resolutions into a bag for the following MIL model [14]. An alternative solution is to construct the patch pyramid to preserve the hierarchical information of WSIs, in which the patches from different resolutions are spatially aligned through the feature pyramid [15]. These multi-resolution methods achieve superior performance than the single-resolution ones. However, these works cannot effectively alleviate the intrinsic semantic gap among different resolution patches, which present different levels of information from cellular-scale to tissue-scale.

Recently, Transformer has been widely used in various vision tasks[16][17][18][19], which can capture the correlation between different segments in a sequence (tokens) to learn the long-range information. Recently, multi-scale vision Transformers (MVTs) have been developed to process the visual tokens of different scales, and can effectively learn and fuse multi-scale features with superior performance for different tasks [20][21][22]. Hierarchical structures are popular in recent MVTs, and their basic idea is to use local Transformer blocks on non-overlapping images and hierarchically aggregate them to learn the multi-scale features [21]. Thus, they can

This work is supported by This work is supported by the National Key R&D Program of China (2021YFA1003004), National Natural Science Foundation of China (62271298, 81871428) and 111 Project (D20031). (Corresponding authors: Jun Shi).

S. Ding, J. Li, J. Wang and J. Shi are with the Key Laboratory of Specialty Fiber Optics and Optical Access Networks, Joint International Research Laboratory of Specialty Fiber Optics and Advanced Communication, Shanghai Institute for Advanced Communication and Data Science, School of Communication and Information Engineering, Shanghai University, China. (Email: junshi@shu.edu.cn).

S. Ying is with the Department of Mathematics, School of Science, Shanghai University, China.

effectively alleviate the intrinsic semantical gap in different scales, and work well on natural images with small image sizes, such as 256×256 and 384×384. However, it is a time-consuming and tedious task to apply existing MVTs to WSIs directly, because WSIs are high-resolution scans of tissue sections, whose full spatial sizes can be over 100000×100000 pixels at 20× magnification.

On the other hand, the spatial information in WSIs can describe the spatial relationship between the tumor and its surrounding tissues, which has great diagnostic significance for WSI analysis [23]. Therefore, it would be much more desirable to learn the spatial-related representation from WSIs for a CAD model. In order to learn spatial-related features, Transformer can add the learnable position encoding to patch embeddings for retaining positional information in the fixed-length sequences [24][25]. However, existing position encoding strategies cannot be used for unfixed-length sequences in the WSI analysis since the number of cropped patches varies among different WSIs. Consequently, the spatial information of WSIs is ignored in the Transformer-based representation learning [17][19], which affects the performance of a CAD model for cancer diagnosis.

Recently, Graph Convolutional Network (GCN) has shown its effectiveness in learning the spatial information of WSIs [23][26][27]. Existing graph-based MIL methods usually regard the WSI as a graph-based data structure, where the nodes correspond to patches and the edges are constructed between adjacent patches [23]. Thus, the constructed graph can effectively represent the spatial relationships among different regions in a WSI. Therefore, it is feasible to introduce the graph representation into Transformer to capture spatial-related information of WSIs.

In this work, we propose a Multi-scale Efficient Graph-Transformer (MEGT) framework to effectively fuse the multi-scale information of WSIs for more accurate diagnosis. Specifically, MEGT adopts a dual-branch Efficient Graph-Transformer (EGT) to process the low-resolution and high-resolution patch embeddings (i.e., tokens in a transformer), respectively. Meanwhile, a multi-scale feature fusion module (MFFM) is developed to fuse these tokens. The proposed EGT integrates the graph representation into the Transformer to capture the spatial information of WSIs. Moreover, to accelerate EGT computation, a novel token pruning module is developed to reduce the redundant tokens. MFFM introduces a non-patch token for each branch as an agent to exchange information with another branch by a specially designed cross-attention. The proposed MEGT is evaluated on two public WSIs datasets, and the results indicate that MEGT outperforms other state-of-the-art (SOTAs) MIL models on the WSI classification task. The main contributions of this work are as follows:

1) A new EGT is proposed to learn both the spatial information of WSIs and the correlations between image patches by innovatively integrating GCN into Transformer. Thus, the EGT can be used as a powerful backbone to provide superior feature representations.
2) A simple yet effective token pruning module is developed in the EGT to reduce the number of tokens without introducing additional parameters, which can significantly expedite EGT computation and preserve the most important tokens.
3) A new MFFM is developed to fuse multi-scale features and alleviate the semantic gap in different resolution patches. The MFFM uses the class token of the branch as an agent to exchange information with the other branch via a cross-attention module, so it only needs linear-time computational and memory complexity.

## II. RELATED WORK

### A. Multiple Instance Learning for WSI

MIL has been successfully applied to WSI analysis, such as cancer diagnosis and survival prediction [28], [29]. According to the network structures, existing MIL methods for WSI analysis generally can be divided into two categories [30]: structure-specific and structure-free models.

The structure-specific model usually adopts the GCN to learn the spatial information from WSI. For example, DeepGraphSurv applied the GCN to WSIs for survival prediction, in which the extracted patches were adopted as nodes for graph construction [23]; Li et al. [31] proposed a context-aware GCN for survival analysis by using the true spatial coordinates of WSIs to connect edges between adjacent patches. These structure-specific models can represent the contextual information between patches for learning the spatial representation of WSIs.

The structure-free model is generally developed based on the attention-MIL. It uses attention scores to weight the instance embeddings to learn the slide-level representation. For example, Li et al. [15] proposed a dual-stream MIL (DSMIL) for WSIs classification, in which the attention scores for each instance were computed with the critical instance; Shao et al. [19] designed a Transformer-based MIL (TransMIL) for WSIs classification, it utilized self-attention to capture long-range dependencies between patches; Huang et al. [17] combined self-supervised learning and Transformer to generate superior feature representation for survival analysis with WSIs. The structure-free model can capture the correlation among different patches to improve the performance of the WSI-based CADs.

Unlike previous MIL methods, we aim to explore a novel Graph-Transformer structure to combine the strengths of the above two models, which integrates GCN into Transformer to learn both the spatial information and the long-range dependencies between image patches in a WSI.

### B. Multi-scale WSI Analysis

In recent years, the multi-scale oriented WSI analysis has attracted more attention. Compared with the single-scale method, the multi-scale approach can learn more semantic information for classification tasks [7] [13][14][15][30]. For example, Campanella et al. [7] trained different MIL branches for different resolutions and then used a max-pooling operation to fuse these multi-resolution embedding for learning the WSI-level representation; Hashimoto et al. [14] mixed patches from different resolutions into a bag and then fed it to the MIL model; Li et al. [15] adopted a pyramidal concatenation strategy to spatially align. patches from different resolutions for WSIs classification; Liu et al. [30] proposed a square pooling layer to align patches from two resolutions, which spatially pooled patches from high-resolution under the guidance of low-resolution; Hou et al. [13] also proposed a heterogeneous graph neural network for tumor typing and staging, in which the heterogeneous graph was constructed based on the feature and spatial-scaling relationship of the multi-resolution patches. These works demonstrate that the multi-scale features can learn more effective slide-level representation to improve the performance of WSI analysis.

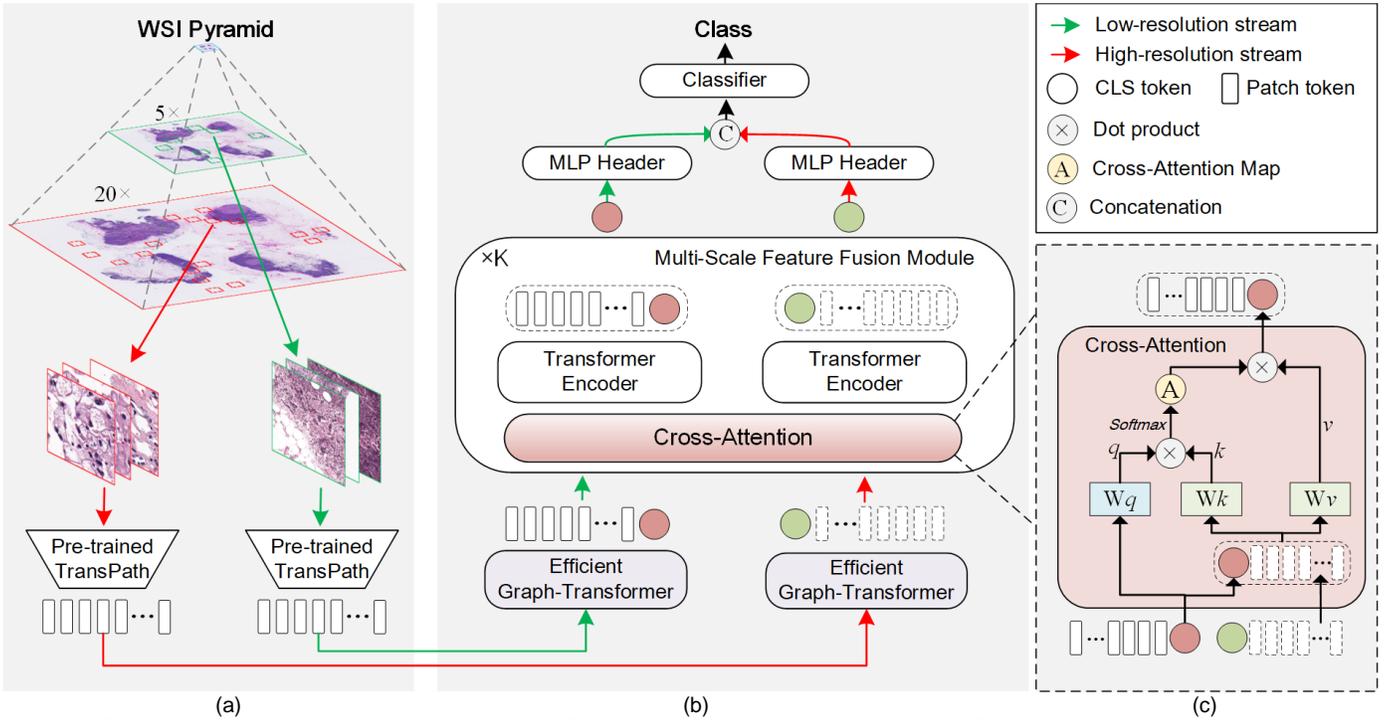

Fig. 1. Overview of MEGT for WSI classification. (a) WSI processing and feature extraction. A WSI pyramid is divided into non-overlapping patches at low and high resolutions, and then their features are extracted by a pre-trained TransPath model, respectively. (b) Flowchart of MEGT. The multi-resolution patch embeddings are fed into the proposed MEGT framework, equipped with the efficient Graph-Transformer and multi-scale feature fusion module, to learn slide-level representation for WSIs classification. (c) Cross-attention operation for the low-resolution branch. The CLS token of the low-resolution branch is used as a query token to interact with the patch tokens from the high-resolution branch through cross-attention, and the high-resolution branch follows the same procedure, but swaps CLS and patch tokens from another branch.

However, due to the semantic gap in different resolution patches, existing multi-resolution schemes still cannot fully utilize the multi-scale features of WSIs. Thus, we suggest to introduce a non-patch agent for each resolution to address this issue.

*C. Multi-scale Vision Transformer*

Inspired by the feature pyramid of images in CNNs, the MVTs have been designed to learn multi-scale visual representations from images [20][21][32][33]. For example, Wang et al. [20] developed a pyramid vision transformer for dense prediction tasks, in which a progressive shrinking pyramid was designed to obtain multi-scale feature maps; Zhang et al. [21] proposed a nested hierarchical Transformer, which stacked Transformer layers on non-overlapping image blocks independently, and then nested them hierarchically. Liu et al. [32] proposed a general Transformer backbone that provided hierarchical feature representations for computer vision; Fan et al. [33] designed a multi-scale vision Transformer for video and image recognition, which hierarchically expanded the feature complexity while reducing visual resolution; These works indicate that the MVTs can learn more effective feature representation for different computer vision tasks.

However, the existing MVTs algorithms are mainly developed for natural images with small sizes. Since WSIs have an extremely large size, these algorithms cannot be directly applied to gigapixel WSIs. Therefore, we investigate how to learn multi-scale feature representations in Transformer models for WSIs classification.

III. METHODOLOGY

In this work, a novel MEGT framework is proposed for WSIs classification, which can effectively exploit the multi-scale feature information of WSIs. Given a WSI pyramid, our framework aggregates patch features of different resolutions to implement the slide-level prediction. As shown in Fig.1, a WSI pyramid is first cropped into non-overlapping patches at low and high resolutions, and their features are extracted by a pre-trained TransPath [34], respectively. Then, the multi-resolution patch embeddings will be fed into the proposed MEGT framework, which consists of two Efficient Graph-based Transformer (EGT) modules and a multi-scale feature fusion module (MFFM), to extract discriminative representation and fully mine multi-scale information. Finally, a WSI-level classifier is employed to generate the slide-level prediction based on the learned representation. In the following subsections, we will introduce the EGT, MFFM, and the learning strategy of the whole framework in detail.

### A. Efficient Graph-Transformer

The spatial correlation among different patches is essential for cancer diagnosis on WSIs. Different from the previous Graph-based and Transformer-based algorithms, our proposed EGT integrates a graph representation of WSI and a Transformer feature to learn both the spatial information of WSI and the long-range dependencies between image patches.

As shown in Fig. 2, the EGT contains two Transformer encoders, a token pruning module, and a Graph-Transformer layer. The first Transformer encoder adopts class token to learn the global information of patch tokens and provides attention scores for token pruning. Then, the token pruning module selects Top-$k$ most important tokens according to the attention scores to reduce the number of tokens. Subsequently, the Graph-Transformer layer uses these selected tokens to simultaneously learn the local and global information of the WSI. Finally, the class token learns all the information again through the second Transformer encoder for the subsequent MFFM.

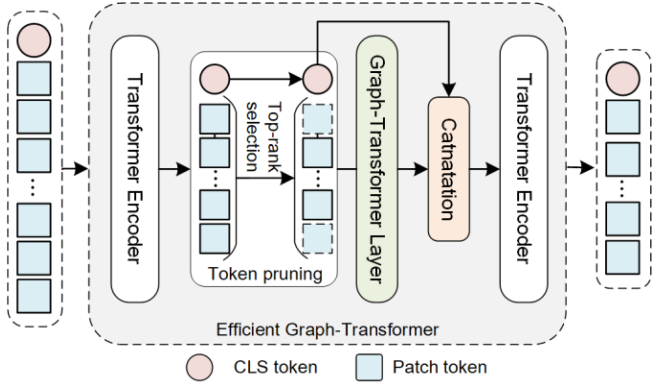

Fig. 2. Structure of Efficient Graph-Transformer (EGT), which is composed of two Transformer encoder layers, a token pruning module, and a Graph-Transformer layer.

#### 1) Transformer Encoder

A Transformer encoder [35] is employed to learn potential long-term dependencies between instances. It contains multiple Transformer layers, each of which has a multi-head self-attention (MSA) and a multi-layer perceptron (MLP). MSA uses the self-attention mechanism to calculate the correlation matrix between instances, and the complexity of memory and time are both $O(n^2)$. In WSI processing, a WSI may be divided into tens of thousands of patches. To address the issue of long instance sequence, we employ the Nystrom-attention (NA) method [36] here, which utilizes the Nystrom method to approximate the standard self-attention. The NA method can be defined as:

$$Q = XW_q, \quad K = XW_k$$
$$NA = softmax\left(\frac{Q\tilde{K}^T}{\sqrt{d}}\right)\left(softmax\left(\frac{\tilde{Q}\tilde{K}^T}{\sqrt{d}}\right)\right)^{\dagger} softmax\left(\frac{\tilde{Q}K^T}{\sqrt{d}}\right) \quad (1)$$

where $W_q$ and $W_k$ are learnable parameters, $Q$ and $K \in \mathbb{R}^{n \times d}$, $d$ is the patch embedding feature dimension, $\tilde{Q}$ and $\tilde{K} \in \mathbb{R}^{m \times d}$ are the $m$ selected landmarks from the $Q$ and $K$, $A^\dagger$ is the Moore-Penrose inverse of $A$. When $m$ is much less than $n$, the computational complexity is reduced from $O(n^2)$ to $O(n)$. The output of the $i$-th Transformer layer can be defined as:

$$T'_i = MSA(LN(T_{i-1})) + T_{i-1}$$
$$T_i = MLP(LN(T'_i)) + T'_i \quad (2)$$

where LN is Layer normalization [37].

#### 2) Token Pruning Module

Although the above NA solves the long-sequence problem in Transformer, the computational complexity is still heavy, when all tokens are used to construct a graph in GCN. Therefore, we perform token pruning to reduce redundant tokens.

Let $n$ denote the number of patch tokens, and an extra class token is added for classification in the Transformer. The interactions between class and other tokens are performed via the attention mechanism in NA, and an attention map $A \in \mathbb{R}^{(n+1) \times (n+1)}$ is obtained, in which the first row of $A$ represents the attention score $a = A[0, 1:] \in \mathbb{R}^{1 \times n}$ from class to all patch tokens. Thus, the attention scores are used to determine the importance of each patch token.

In the multi-head self-attention layer, there are multiple class attention vectors $a^h$, where $x\ h = [1, \ldots, H]$, and $H$ is the total number of attention heads. We compute the average value of all heads by:

$$\bar{a} = \sum_{h=1}^{H} a_h / H \quad (3)$$

After that, we select the tokens corresponding to the $k$ largest (top-$k$) elements in $\bar{a}$, and further fuse the other tokens into a new token using the attention scores in $\bar{a}$. Therefore, the token fusion can be written as:

$$h_{fusion} = \sum_{i=1}^{i=n-k} h_i \bar{a}_i \quad (4)$$

where $h_i$ represent the $i$-th patch token.

#### 3) Graph-Transformer Layer

After token pruning, a total $(k + 1)$ patch tokens are used in the Graph-Transformer layer. Note that we do not use class token, since it may affect the learning of GCN. Furthermore, graph construction is an essential step in the Graph-Transformer layer for graph representation learning. A graph can be denoted as $G = (V, E)$, where $V$ and $E$ represent the nodes and a set of edges in the graph, respectively. We take each patch token as a node, and a bag can be represented as a graph node feature matrix $G_{in} \in \mathbb{R}^{(k+1) \times d}$, which has $(k + 1)$ nodes, and each node has a $d$-dimensional feature vector.

Instead of using the $k$-NN algorithm to construct the adjacency matrix $A_{adj} \in \mathbb{R}^{(k+1) \times (k+1)}$ based on a pair of node features, we adopt the attention matrix of self-attention to adaptively generate the adjacency matrix, thereby further speeding up the training of the network. As shown in Fig. 3, the Graph-Transformer layer is similar to the self-attention that maps queries and key-value pairs to outputs. The difference is that the Graph-Transformer layer adds a branch to perform graph convolution.

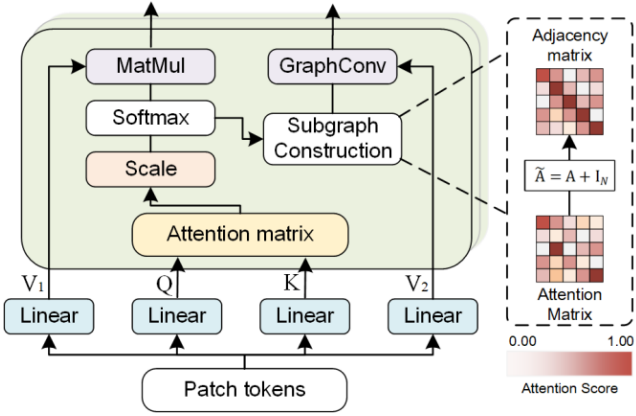

Fig. 3. Structure of Graph-Transformer layer.

Given an input patch embedding $X_{patch} \in \mathbb{R}^{(k+1) \times d_k}$, the matrices of query $Q$, key $K$ and values $V_1, V_2$ are first calculated through four different linear projections as follows:

$$\begin{aligned} Q &= Liner(X_{patch}) = X_{patch}W_Q, \\ K &= Liner(X_{patch}) = X_{patch}W_K, \\ V_1 &= Liner(X_{patch}) = X_{patch}W_{V_1}, \\ V_2 &= Liner(X_{patch}) = X_{patch}W_{V_2}. \end{aligned} \quad (5)$$

where $W_Q, W_K, W_{V_1}$, and $W_{V_2} \in \mathbb{R}^{d_k \times d_m}$ are the corresponding weight matrices of linear projections.

In addition, we also adopt the multi-head structure to expand the Graph-Transformer layer. As shown in Fig. 3, this structural design can project the inputs into different subspaces to learn different features, thereby improving the performance of the model. Specifically, the input features are evenly split into $h$ parts, and the attention matrix can be calculated as follows:

$$A_i = Score(Q_i, K_i) = \frac{Q_i K_i^T}{\sqrt{d_m/h}} \quad (6)$$

where $A_i \in \mathbb{R}^{(k+1) \times (k+1)}, i = [1, \ldots, h]$, $Q_i \in \mathbb{R}^{(k+1) \times \frac{d_m}{h}}, K_i \in \mathbb{R}^{(k+1) \times \frac{d_m}{h}}$, and $1/\sqrt{d_m/h}$ is a scaling factor.

For the Transformer branch, the outputs of the multi-head structure are first concatenated together and then fed into linear projections to obtain the complete outputs:

$$\begin{aligned} V'_{1i} &= softmax(A_i)V_{1i} \\ V'_1 &= Concat(V'_{11}, \ldots V'_{1h})W_{o1}. \end{aligned} \quad (7)$$

where $V'_{1i} \in \mathbb{R}^{(k+1) \times \frac{d_m}{h}}, V'_1 \in \mathbb{R}^{(k+1) \times d_k}$, and $W_{o1} \in \mathbb{R}^{d_m \times d_k}$ the weight matrices of linear projection.

For the GCN branch, the adjacency matrix $\widetilde{A}_i$ is first transformed by the normalized $A_i$ with self-connections, and then the $\widetilde{A}_i \in \mathbb{R}^{(k+1) \times (k+1)}$ and node embedding $V_{2i} \in \mathbb{R}^{(k+1) \times \frac{d_m}{h}}$ are fed into the GCN [38] to learn graph representations. The forward propagation of GCN can be written as follows:

$$V'_{2i} = \sigma\left(\widetilde{D}^{-\frac{1}{2}}\widetilde{A}\widetilde{D}^{-\frac{1}{2}}V_{2i}W^{(l)}\right) \quad (8)$$

where $V'_{2i} \in \mathbb{R}^{(k+1) \times \frac{d_m}{h}}$, $\widetilde{D}$ is the degree matrix of $\widetilde{A}$, and $W^{(l)}$ is trainable weight matrix. After that, the multiple subgraph representations are concatenated together and then fed into the linear projections to obtain the complete outputs:

$$V'_2 = Concat(V'_{21}, \ldots V'_{2h})W_{o2} \quad (9)$$

where $V'_2 \in \mathbb{R}^{(k+1) \times d_k}$ and $W_{o2} \in \mathbb{R}^{d_m \times d_k}$ are the weight matrices of linear projection.

The adjacency matrix can effectively represent the spatial distribution and adjacent relationship between nodes, and then the spatial-related information is learned through GCN layer. Thus, the generated graph representation contains the local information and short-range structure, which are ignored in the original Transformer.

Finally, the Transformer branch output $V'_1$ and GCN output $V'_2$ are concatenated together and then fed into linear projections to fuse the local and global information:

$$X'_{patch} = Concat(V'_1, V'_2)W_{o3} \quad (10)$$

where $X'_{patch} \in \mathbb{R}^{(k+1) \times d_k}$ and $W_{o3} \in \mathbb{R}^{2d_k \times d_k}$ are the weight matrices of linear projection.

### B. Multi-scale Feature Fusion Module

As mentioned before, the patches with different resolutions present different diagnostic information ranging from cellular-scale (e.g., nucleus and micro-environment) to tissue-scale (e.g., vessels and glands), which has an intrinsic semantical gap. However, existing multi-scale methods have not paid enough attention to this issue. To this end, we propose a novel MFFM, which alleviates the semantical gap by using the class token as an agent to exchange information between two resolutions.

As shown in Fig. 1(b), MEGT contains $K$ MFFM, each of which consists of two Transformer encoders and an efficiently cross-attention, and the value of $K$ is set to 2 in this work. Compared with CNN, vision Transformer adds a learnable class token to summarize all patch tokens. Since the class token uniformly converts different resolution features into discriminative representation for classification, we use class tokens to effectively fuse features of different scales via a Cross-Attention (CA). Fig. 1(c) shows the basic idea of CA, which uses the class token of a branch to exchange information with patch tokens of the other branch. Since the class token already learns the global information of all patch tokens in the corresponding branch, interacting with patch tokens of another branch can help to learn more information at different scales. After the CA, the class token passes the learned information to its patch tokens at the later Transformer encoder, thereby enriching the representation of patch tokens.

Let $X^i = [X^i_{cls}|X^i_{patch}]$ be the token sequence at branch $i \in$ [low, high], where $X^i_{cls}$ and $X^i_{patch}$ represent the class and patch token of branch $i$, respectively. We first collect $X^{high}_{patch}$ from high-branch, and then concatenate it with $X^{low}_{cls}$ to obtain the $X' = [X^{low}_{cls} \| X^{high}_{patch}]$. Here, $X^{low}_{cls}$ is used as the query and $X'$ works as the key and value to perform the CA operation, which can be expressed as:

$$Q = X^{low}_{cls} W_q, \ K = X' W_k, \ V = X' W_v$$
$$CA = softmax\left(\frac{QK^T}{\sqrt{d}}\right)V \quad (11)$$

where the $X^{low}_{cls} \in \mathbb{R}^{1 \times d}$, so the computational complexity of CA is $O(n+1)$ instead of $O(n^2)$ for self-attention, $n$ is the number of patch tokens in the high-branch. The high-branch follows the same procedure, but $X^{high}_{cls}$ is used as the query and $X' = [X^{high}_{cls} \| X^{low}_{patch}]$ is used as the key and value. The output $O^{low}$ with the CA operation can be expressed as:

$$y^{low}_{cls} = X^{low}_{cls} + CA(X^{low}_{cls}, X')$$
$$O^{low} = [y^{low}_{cls} \| X^{low}_{patch}] \quad (12)$$

### C. Network Architecture and Training Strategy

EGT and MFFM are the basic components of the proposed MEGT. As shown in Fig. 1(b), MEGT contains two separate branches, and EGT is used in each branch to provide superior feature representation for the MFFM. Then, MFFM employs the class tokens to fuse multi-scale features multiple times via a cross-attention layer. Finally, the dual-resolution class tokens are concatenated to produce slide-level representation $Z$ for WSIs classification, which can be defined by:

$$\begin{cases} Z = concat(X^{low}_{cls}, X^{high}_{cls}) \\ \widehat{Y} = softmax(MLP(Z)) \end{cases} \quad (13)$$

For the model training, the loss function $\mathcal{L}$ is defined by the cross entropy between the bag class labels $Y$ and bag class predictions $\widehat{Y}$ which can be expressed as:

$$\mathcal{L} = -\frac{1}{M}\sum_{i=1}^{M}\sum_{j=1}^{C} Y_{ij} \log(\widehat{Y}_{ij}) \quad (14)$$

where $M$ is the number of patients, $C$ is the number of classes. The gradient descent algorithm is adopted for optimizing the whole model.

## IV. EXPERIMENTS AND RESULTS

### A. Datasets

The proposed MEGT was evaluated on two commonly used WSI datasets, namely the Cancer Genome Atlas Renal Cell Carcinoma (TCGA-RCC) dataset (https://portal.gdc.cancer.gov) and CAMELYON16 dataset [39].

TCGA-RCC is a WSI dataset for Renal Cell Carcinoma classification, and it contains three categories: Kidney Chromophobe Renal Cell Carcinoma (KICH), Kidney Renal Clear Cell Carcinoma (KIRC), and Kidney Renal Papillary Cell Carcinoma (KIRP). A total of 914 slides are collected from 876 cases, including 111 KICH slides from 99 cases, 519 KIRC slides from 513 cases, and 284 KIRP slides from 264 cases. After WSI pre-processing, the mean numbers of patches extracted on each slide at 5× and 20× magnification were 4263 and 14627, respectively.

CAMELYON16 is a public dataset for metastasis detection in breast cancer, including 270 training and 129 test slides. After WSI pre-processing, there were about mean 918 and 3506 patches selected from each slide at 5× and 20× magnifications, respectively.

### B. Experiment Setup

In our experiments, all algorithms were trained on the 270 official training slides and tested on the 130 official test slides in the CAMELYON16 dataset. These training slides were further randomly divided into training and validation sets with a ratio of 9:1. Meanwhile, the results of the CAMELYON16 dataset were obtained on the official testing set. For the TCGA-RCC dataset, we conducted a 5-fold cross-validation on the 936 slides to evaluate these algorithms. The widely used accuracy, recall, F1-score (F1), and area under the curve (AUC) were used as evaluation indices to compare the classification performance of different algorithms. The results of the TCGA-RCC dataset were presented in the format of mean ± SD (standard deviation). Since the classification algorithms consistently achieved better results on 20× images than those on 5× ones, we only reported the experimental results on a single 20× scale for both datasets.

### C. Implementation Details

In WSI pre-processing, each WSI was divided into non-overlapping 299×299 patches in both the magnifications of 20× and 5×, and a threshold was set to filter out background patches. After patching, we used the pre-trained TransPath [34] model, a pre-training vision Transformer for histopathology images, to extract a feature vector with a dimensional of 768 from each 299×299 patch. Thereafter, for the proposed MEGT, the Adam optimizer was used in the training stage with a learning rate of 1e-4 with a weight decay of 1e-5, to update the models. The size of the mini-batch was set as 1 (bag). The MEGT and other models were trained for 150 epochs with a cross-entropy loss function, and they would early stop if the loss would not decrease in the past 30 epochs. All models were implemented by Python 3.7 with PyTorch toolkit 1.10.0 on a platform equipped with one NVIDIA GeForce RTX 3090 GPU.

### D. Comparison with State-of-the-art Methods

We compared the proposed MEGT with the following SOTA MIL algorithms:
1) The conventional MIL with the pooling operators, including Mean-pooling and Max-pooling.
2) The classic attention-based pooling operator ABMIL [6] and its variant, non-local attention pooling DSMIL [15].
3) The Transformer-based MIL, TransMIL [19].
4) The Cluster-based MIL, Re-Mix [40]
5) The Graph-based MIL, GCN-MIL [23], and GTP [41].
6) The multi-resolution MIL approaches, including MS-Max [7], MS-ABMIL [14], DSMIL-LC [15] and H²-MIL [13].

Table I shows the overall classification results of the comparison experiment on the TCGA-RCC and CAMELYON16 datasets. Generally, the multi-scale MIL algorithms achieve better results than the single-scale MIL ones, which indicates that the multi-scale information of WSIs is important for cancer diagnosis. Moreover, the proposed MEGT achieves the best mean accuracy of 96.91±1.24%, recall of 97.65±0.86%, and F1-score of 96.26±1.19% on the TCGA-RCC dataset. Meanwhile, compared to other algorithms, it improves at least 1.47%, 1.54%, and 1.37% on the corresponding indices, respectively. Similarly, MEGT outperforms all the compared algorithms with the best accuracy of 96.89%, F1-score of 95.74%, and AUC of 97.30% on the CAMELYON16 dataset. As a SOTA multi-resolution MIL algorithm, $H^2$-MIL achieves the second-best performance due to a newly proposed GCN algorithm in it, which can learn hierarchical representation from a heterogeneous graph. Nevertheless, our MEGT still improves by 0.77%, 1.00%, and 0.60% on classification accuracy, recall and F1-score, respectively, over the $H^2$-MIL.

On the other hand, the proposed EGT outperforms all the other single-scale MIL algorithms on all indices for the single-scale experiment on the single 20× scale images. On the TCGA-RCC dataset, EGT achieves the best classification accuracy of 96.91%, recall of 97.65%, and F1-score of 96.26%, respectively. Meanwhile, compared to other algorithms, it improves at least 1.17%, 1.54% and 1.37% on the corresponding indices, respectively. EGT also gets a similar trend on the CAMELYON16 dataset with the best classification performance of 96.12%, 94.85%, and 96.34% on the accuracy, F1-score and AUC, respectively. Compared to other algorithms, it improves at least 0.77%, 1.45%, and 0.55%, on the corresponding indices respectively.

### E. Ablation Study

To further evaluate the effectiveness of MEGT, we conducted a series of ablation experiments to delineate the contributions of two major components in the proposed MEGT: the EGT and the MFFM.

#### 1) Effects of Efficient Graph-Transformer

To evaluate the effectiveness of the proposed token pruning module (TPM) and Graph-Transformer Layer (GTL) in EGT, we compared the proposed EGT with its three variants:
1) EGT-m: This variant removed both the TPM and GTL, which was then equivalent to TransMIL.
2) EGT-TPM: This variant only maintained the TPM, but removed the GTL.
3) EGT-GTL: This variant only maintained the GTL, but removed the TPM.
4) EGT-KNN: This variant used the same structure as the EGT, but used the KNN algorithm to perform the subgraph construction.

Fig. 3 shows the results of different variants on both the TCGA-RCC and CAMELYON16 datasets. It can be seen that both EGT-TPM and EGT-GTL outperform EGT-m, suggesting the effectiveness of TPM and GTL in EGT. Moreover, TPM can effectively reduce the number of redundant tokens without additional parameters, and GTL can learn more important local-global features with lower computational complexity. Therefore, by integrating TPM and GTL, EGT can efficiently learn both the spatial-related information of WSI and the correlation between patch tokens for improved performance. In addition, although the EGT-KNN achieves similar performance to our EGT, it needs more time to construct the adjacency matrix, resulting in a significant decrease in the training speed of the network.

TABLE I
COMPARISON RESULTS ON THE CAMELYON16 AND TCGA DATASETS. WE USE THE SAME PATCH FEATURE EXTRACTOR FOR ALL METHODS (UNIT: %)

| Method | Resolution | TCGA-RCC | | | CAMELYON16 | | |
|---|---|---|---|---|---|---|---|
| | | ACC | Recall | F1 | ACC | F1 | AUC |
| Mean-pooling | 20× | 89.67±1.06 | 90.74±1.46 | 88.66±1.24 | 91.47 | 87.64 | 92.75 |
| Max-pooling | 20× | 92.20±0.72 | 92.51±0.81 | 91.01±0.90 | 92.25 | 88.64 | 93.64 |
| ABMIL [6] | 20× | 93.72±1.31 | 94.53±1.31 | 92.53±1.83 | 93.80 | 90.11 | 94.82 |
| GCN-MIL [23] | 20× | 93.28±1.10 | 93.45±1.15 | 91.90±1.29 | 93.02 | 89.89 | 94.17 |
| DSMIL [15] | 20× | 92.90±0.84 | 93.27±0.69 | 91.95±1.21 | 93.80 | 91.30 | 94.94 |
| DSMIL* [15] | 20× | - | - | - | 86.82 | - | 89.44 |
| TransMIL [19] | 20× | 93.94±1.09 | 94.87±0.64 | 93.43±0.75 | 94.57 | 92.93 | 95.53 |
| TransMIL* [19] | 20× | - | - | - | 88.37 | - | 93.09 |
| Re-Mix [40] | 20× | 93.58±1.25 | 94.37±0.95 | 93.01±0.84 | 94.57 | 93.48 | 96.03 |
| Re-Mix* [40] | 20× | - | - | - | 95.35 | 95.18 | - |
| GPT [41] | 20× | 94.27±1.12 | 94.75±1.42 | 93.44±1.58 | 95.35 | 93.75 | 95.79 |
| **EGT (Ours)** | 20× | **95.37±0.64** | **96.04±0.40** | **94.73±0.55** | **96.12** | **95.20** | **96.34** |
| MS-Max-pooling [7] | 20×+5× | 91.42±0.98 | 91.98±0.79 | 90.68±1.14 | 93.02 | 90.53 | 94.26 |
| MS-ABMIL [14] | 20×+5× | 94.58±0.87 | 95.43±1.45 | 94.08±1.31 | 94.57 | 92.47 | 95.15 |
| DSMIL-LC* [15] | 20×+5× | - | - | - | 89.92 | - | 91.65 |
| $H^2$-MIL [13] | 20×+5× | 95.44±0.96 | 96.11±1.02 | 94.89±1.34 | 96.12 | 94.74 | 96.70 |
| **MEGT (Ours)** | 20×+5× | **96.91±1.24** | **97.65±0.86** | **96.26±1.19** | **96.89** | **95.74** | **97.30** |

* DENOTES SCORES REPORTED IN THE PAPER.

TABLE II
ABLATION STUDY RESULTS FOR EVALUATING THE TPM AND GTL IN EGT ON THE CAMELYON16 AND TCGA DATASETS (UNIT: %)

| Method | TPM | GTL | TCGA-RCC ACC | TCGA-RCC F1 | CAMELYON16 ACC | CAMELYON16 AUC | Average Seconds (Epoch) TCGA-RCC | Average Seconds (Epoch) Came.16 | Model Size |
|---|---|---|---|---|---|---|---|---|---|
| EGT-m | × | × | 93.83±1.11 | 92.65±1.09 | 93.80 | 94.72 | 42s | 27s | 2.5 M |
| EGT-TPM | √ | × | 94.41±1.32 | 92.88±1.28 | 94.57 | 95.55 | 29s | 18s | 2.5 M |
| EGT-GTL | × | √ | 94.58±1.03 | 93.24±1.11 | 95.35 | 95.69 | 65s | 39s | 4.2 M |
| EGT-KNN | √ | √ | 95.21±0.98 | 94.53±1.24 | 96.12 | 96.59 | 141s | 101s | 4.2 M |
| EGT | √ | √ | 95.37±0.64 | 94.73±0.55 | 96.89 | 97.30 | 38s | 26s | 4.2 M |

*2) Effects of Multi-scale Feature Fusion Module*

We further compared our cross-attention fusion strategy to several other feasible multi-scale fusion methods, including:
1) Concatenation: It removed the cross-attention in the MFFM, and then only concatenated the class tokens of the two branches after MFFM.
2) All-attention: It concatenated all tokens from different branches together and then fused them via the MSA.
3) Class-Token: It simply averaged the class tokens on the two branches, so that the information was passed back to patch tokens through the later transformer encoder.

Fig. 4 shows the classification results of four different token fusion strategies on the TCGA-RCC and CAMELYON16 datasets. The specially designed cross-attention outperforms all the other strategies on all indices, which indicates it can effectively fuse multi-scale information of WSIs with superior performance. It is worth noting that although the All-attention mechanism uses additional self-attention to learn information between two branches, it fails to achieve better performance compared to the simple Class-Token. The experimental results demonstrate that class tokens can avoid the semantic gap in different resolution patches, resulting in the performance improvement of the MFFM.

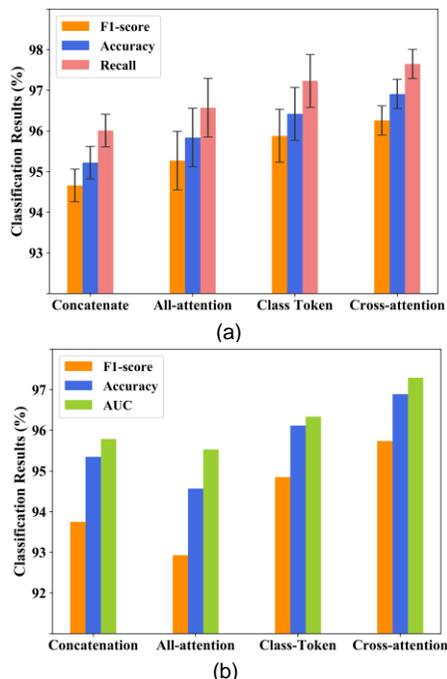

Fig. 4. Classification results for evaluating different fusion strategies in MFFM on (a) TCGA-RCC dataset and (b) CAMELYON16 dataset.

*F. Visualization of Attention Weights*

We further used the CAMELYON16 test set with pixel-level annotations to evaluate the ability of our MEGT to locate the positive instances. For the global attention heatmaps in the second column of Fig. 5, the attention weights are normalized between 0 to 1 (i.e., blue to red) in each cross-attention map. Thus, the red regions in the global attention heatmaps represent the highest contribution instances for classification in each bag. It can be seen that the hot regions predicted by our model tend to appear in the annotated ROIs on both WSIs. Moreover, almost all the high-attention patches contain high-density irregular cells, which proves that our framework can effectively focus on the most discriminative patches via the cross-attention module to implement a more accurate WSIs classification.

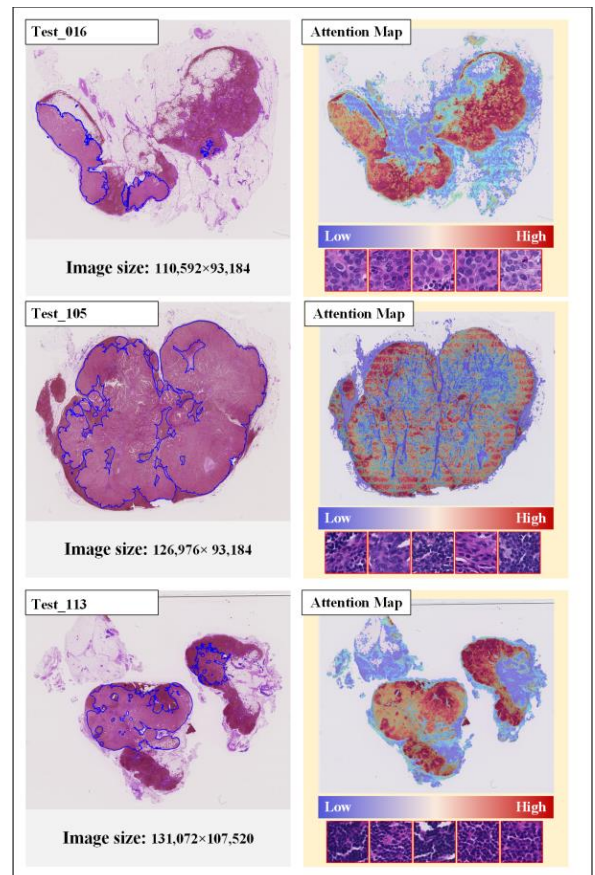

Fig. 5. Visualization of cross-attention maps on Camelyon16 testing set. Each representative slide is annotated by pathologists, who roughly highlighted the tumor tissue regions (left). A global attention map with corresponding high-attention patches to each slide is generated by computing the attention weights of cross-attention maps (right).

## G. Parameter Sensitivity Analysis

A parameter sensitivity analysis was also conducted for the proposed MEGT. Two architecture parameters in MEGT will affect the classification performance, i.e., the number of Transformer layers $L$ in MFFM and the number of MFFM $K$. We then changed both parameters to investigate their impact on the WSI classification task. The five models (A-E) in Table III represent different combinations of architecture parameters $L$ and $K$.

Table III shows the classification results of different numbers of Transformer layers on the low-branch and high-branch. It can be found that both models A and C significantly increase parameters but without any improvement in accuracy compared to the original MEGT, because more Transformer layers lead to a larger number of parameters, which may suffer from the overfitting problem. It is worth noting that the performance of MEGT will be decreased by reducing the depth of the high-branch in model B, which indicates that the high-branch plays the main role in learning the features of WSIs, while the low-branch only provides additional information.

The number of MFFM $K$ is an important parameter in our MEGT, which controls the fusion frequency of the two branches. Table III shows the classification results with different numbers of MFFMs on the TCGA-RCC dataset and CAMELYON16 dataset. With MEGT as the baseline, the accuracies of the models D and E on both datasets are much degraded by using only one MFFM, because the class token cannot pass the learned information to its patch tokens. In addition, too much branch fusion does not improve performance, but introduces more parameters. This is because the cross-attention is a linear operation without any nonlinearity function, which results in overfitting of the model due to over-parameterization. Considering the performance and parameters of the model, we finally select 1, 2 and 2 as the values of transformer layers $L_{low}$, $L_{high}$ and MFFM $K$.

TABLE III
CLASSIFICATION RESULTS WITH DIFFERENT ARCHITECTURE PARAMETERS ON CAMELYON16 AND TCGA-RCC DATASETS

| Model | L | | K | Came16 Acc. (%) | TCGA Acc. (%) | Params. (M) |
|---|---|---|---|---|---|---|
| | l | h | | | | |
| GCMT | 1 | 2 | 2 | 96.89 | 96.91±1.24 | 26.79 |
| A | **2** | 2 | 2 | 96.12 | 96.21±1.13 | 33.09 |
| B | 1 | **1** | 2 | 95.35 | 95.98±1.38 | 20.49 |
| C | 1 | **4** | 2 | 96.12 | 96.67±1.26 | 39.39 |
| D | 1 | 2 | **1** | 94.57 | 95.77±1.85 | 12.61 |
| E | 1 | 2 | **4** | 95.35 | 96.79±1.47 | 48.85 |

THE **BLACK** COLOR INDICATES CHANGES FROM MEGT.

## V. DISCUSSION

In this work, a novel MEGT network is proposed for cancer diagnosis on gigapixel WSIs. MEGT designs an effective MFFM to aggregate different resolution patches to produce stronger WSI-level representation, which is easy to implement without complicated patch processing. Experimental results on both TCGA-RCC and CAMELYON16 datasets indicate the effectiveness of our proposed MEGT.

Previous works on WSIs generally focused on single-resolution methods, which fail to capture multi-scale information of WSIs. Inspired by the diagnosis process of pathologists, some researchers have extended previous MIL algorithms to learn multi-scale representations from the WSI pyramid. From the aspect of multi-scale feature fusion, existing schemes are restricted to simple concatenation or multi-scale feature pyramid construction, which are not paid enough attention to the intrinsic semantic gap among different resolution patches. To this end, our proposed MEGT introduces a class token for each resolution as an agent to fuse multi-scale information of WSIs. Since the class token uniformly converts different resolution features into slide-level representations for classification, the semantic gap in different resolution patches is alleviated. In addition, our framework avoids the complicated patch processing, such as building feature pyramids and heterogeneous graphs in the WSI. Therefore, the proposed MEGT is a simple yet effective framework for learning the multi-scale information of WSIs for a CAD model.

The spatial information of WSIs is also essential for cancer diagnosis. Different from previous graph-based MIL or Transformer-based MIL, the proposed EGT aims to learn the spatial features from graph data to enhance the performance of the Transformer. Here, each node in the graph data corresponds to a patch in the original WSI, and the edges are computed by the embedded features from these patches. Thus, the patch-based graph actually represents the spatial relationships among different regions in a WSI. In addition, EGT utilizes the attention scores of Transformer encoder for token pruning, which greatly reduces the computational complexity of graph construction and graph convolution. Therefore, EGT can efficiently learn the spatial information and the correlation between patches simultaneously to produce a superior feature representation.

Although the proposed MEGT achieves superior performance over the compared SOTA algorithms, it still has some room for improvement. For example, we will focus on the data-driven pretext task design in self-supervised learning to learn more effective multi-scale feature representation, which can alleviate the issue of a small sample size in histopathological images. On the other hand, MEGT is currently only suitable for dual resolution on WSIs due to the multi-scale feature fusion strategy of MFFM. Future studies can explore other efficient strategies, such as feature pyramids and hierarchical networks, to combine more resolution for feature fusion.

## VI. CONCLUSION

In this work, we proposed Multi-scale Efficient Graph-Transformer (MEGT), a dual-branch Transformer for aggregating image patches of different resolutions, to promote the accuracy of cancer diagnosis on WSIs. Particularly, an effective MFFM was developed to learn the multi-scale features and reduce the semantic gap in different resolution patches. Meanwhile, the EGT was specifically designed to improve the ability of the branches in MEGT for learning spatial information of WSIs. Experimental results on two public WSI datasets demonstrated the effectiveness of the proposed MEGT framework. It suggests that MEGT has the potential for WSI-based CAD in clinical practice.